\documentclass{article}
\usepackage{iclr2027_conference,times}
\iclrfinalcopy

\usepackage{amsmath,amsfonts,bm}

\def\eqref#1{equation~\ref{#1}}

\def\1{\bm{1}}

\DeclareMathAlphabet{\mathsfit}{\encodingdefault}{\sfdefault}{m}{sl}
\SetMathAlphabet{\mathsfit}{bold}{\encodingdefault}{\sfdefault}{bx}{n}

\usepackage{amsmath,amssymb,amsthm,booktabs,graphicx,microtype,tabularx,enumitem,xcolor,hyperref,url,placeins,longtable}
\hypersetup{colorlinks=true,linkcolor=black,citecolor=black,urlcolor=black,
pdftitle={PairAudit: Guiding Human Review with Graph Tokens under Distribution Shift},
pdfauthor={Jiran Tao and Binyan Jiang}}
\newtheorem{proposition}{Proposition}
\newcommand{\method}{\textsc{PairAudit}}

\newcommand{\ind}{\mathbf{1}}
\title{\method: Guiding Human Review with Graph Tokens under Distribution Shift}
\author{Jiran Tao \quad Binyan Jiang\\
The Hong Kong Polytechnic University\\
\texttt{22117758r@connect.polyu.hk} \quad \texttt{by.jiang@polyu.edu.hk}}
\begin{document}
\maketitle
\lhead{Preprint}
\begin{abstract}
Intrusion detectors can confidently misclassify attacks that were not seen during training. Human review can correct these errors, but only a limited number of cases can be checked. Uncertainty-based review may overlook confident errors, while anomaly scores alone do not show whether changing the review plan will correct more errors. We introduce \method\ to find overlooked errors and improve review under a fixed budget. Its graph tokens capture prediction patterns across connected nodes. Rather than building another predictor through feature aggregation, \method\ uses unusual relational patterns to uncover potential errors in existing predictions. Human feedback then helps decide whether these findings justify changing review priorities. Experiments across security tasks show that \method\ corrects more errors on average than uncertainty-based review, including more errors on unseen attacks. These gains account for all review costs and do not require retraining the detector.
\end{abstract}

\section{Introduction}
A detector can work well on familiar cases and still make confident mistakes when new patterns appear. This is a practical problem under out-of-distribution (OOD) shift. In security, an unseen attack subtype may be mistaken for benign activity even though its correct label still belongs to a known category. Human review can correct these mistakes while the detector stays in use. The difficulty is that reviewers can check only a small part of the data, so errors outside their review queue may remain unnoticed.

A natural approach is to review the cases the detector is least certain about. This makes good use of its predictions, but can miss errors made with high confidence. OOD and error-detection scores offer other ways to order cases \citep{hendrycks2017baseline,liu2020energy,jiang2018trust,wu2023gnnsafe,ma2024revisiting}. Yet a different ranking is not automatically a better review plan. Bringing one case into a fixed budget means leaving another out. A useful method must find errors the current queue overlooks and help decide whether the proposed change is worth making.

Graph structure offers a way to find such cases. A node may look ordinary on its own while its prediction forms an unusual pattern with those of its neighbors. But disagreement alone is not enough: some kinds of nodes routinely differ from their neighbors. For example, a legitimate shared server may connect to many devices flagged as malicious. Its own benign prediction can still be correct, and such disagreement may be familiar for servers with that role. Treating these familiar patterns as suspicious can waste review effort. We therefore ask whether the observed relationship is expected for a node with those features, rather than treating every unusual-looking relationship in the same way.

We introduce \method\ to connect this relational evidence with decisions about human review. Its graph tokens describe the predictions of a node and its neighborhood. A density model learns several familiar patterns within each known class, and node features determine how much weight each pattern receives. A strong disagreement can be unsurprising when it fits a familiar pattern for those features; a different relationship can still be flagged. The score is fitted and selected using known training and validation labels, without requiring OOD labels.

Uncertainty review can remain useful even when it misses confident errors. \method\ therefore retains part of the original queue (Raw) and proposes replacements from outside its budget. Reviewers provide ordinary class labels, not OOD status or new subtype names. A sequential pair audit uses these labels to decide whether to adopt the remaining exchange. Retention, auditing and subsequent selection form one review policy. Both audited endpoints count toward the same budget. The objective is to correct more errors than the original queue after paying for the decision.

Our contributions are:
\begin{itemize}[leftmargin=*,itemsep=2pt,topsep=3pt]
\item \textbf{Relational evidence for overlooked errors.} We use graph tokens and feature-conditioned pattern densities to guide review of a frozen detector. This accounts for different familiar neighborhood patterns within a class and uses them to assess new cases, without retraining the detector or changing the review labels.
\item \textbf{An adaptive review policy with explicit costs.} We combine Raw retention and sequential pair audits with exact budget and net gain identities. These identities account for both useful audit labels and the reviews they displace. We also characterize uniform pair sampling without assuming independent graph nodes, while keeping these guarantees distinct from the working posterior used for decisions.
\item \textbf{Consistent benefits across the tested attack shifts.} All four primary tasks show positive mean overall and unseen-attack correction gains with high authorization rates. Decisions use about 5\% of the budget on average. Additional tests show gains at low and high OOD prevalence and in multiclass review. The low-OOD test also corrects more errors on known cases. Controls assess robustness to pairing and exchange size and show early rejection of a harmful candidate.
\end{itemize}

\section{Related Work}
\paragraph{OOD scores and review decisions.}
Maximum softmax probability and Energy provide uncertainty and OOD scores \citep{hendrycks2017baseline,liu2020energy}. GNNSafe propagates energy scores over a graph, and GRASP studies when score propagation helps and augments it accordingly \citep{wu2023gnnsafe,ma2024revisiting}. Trust Score assesses predicted-class support through reference distances and explicitly motivates human review prioritization \citep{jiang2018trust}. These methods can induce review rankings. Their scoring procedures, however, do not supply a sequential decision based on current analyst labels about whether to replace part of an existing queue under a shared budget. Their scores alone do not establish a positive correction gain on a new workload. Our policy retains part of Raw and uses paid feedback to guide replacements. Table 4 compares complete review policies at equal capacity. Table~\ref{tab:external} documents how their correction benefits vary across the tested tasks.

\paragraph{Conditional anomaly and graph models.}
Conditional anomaly detection distinguishes unusual behavior from unusual context \citep{song2007conditional}. Conditional mixture densities and graph mixture density networks provide foundations for multimodal modeling \citep{bishop1994mixture,errica2021graph}; graph neighborhood reconstruction exploits relational evidence \citep{roy2024gadnr}. We do not claim these as new primitives. Our graph tokens and feature-conditioned densities model the relational behavior of a frozen classifier and supply proposals whose correction value is assessed through paid review.

\paragraph{Human feedback and allocation.}
Active testing allocates labels for model evaluation \citep{kossen2021active}; human feedback can improve OOD decisions \citep{vishwakarma2024taming,bai2025human}; conservative policy improvement considers retaining a baseline with limited evidence \citep{laroche2019spibb}. Our feedback is the analyst's ordinary class judgment. Its acquisition both corrects an object and consumes capacity that could have corrected another. Consequently, a favorable candidate score or a favorable remaining exchange does not by itself imply a beneficial complete workflow.

\section{Review Objective and Available Information}

The target pool $\mathcal U=\{1,\ldots,N\}$ contains $N$ nodes and $K$ known review classes (e.g., benign and attack for $K=2$). A frozen classifier returns $p_i=(p_{i1},\ldots,p_{iK})$: $p_{ik}$ is its probability for class $k$, with $\sum_kp_{ik}=1$. Its predicted label is $\hat y_i=\arg\max_kp_{ik}$. Reviewers provide $y_i\in\{1,\ldots,K\}$.

Assuming equal review cost per node, let $H<N$ be the maximum number of distinct nodes that the available human budget can cover. It limits how many samples reviewers can see and label. Define the Raw uncertainty score and baseline set by
\begin{equation*}
s_R(i)=1-\max_kp_{ik},\qquad
R_H=\{R(1),\ldots,R(H)\}.
\end{equation*}
Here $R=(R(1),\ldots,R(N))$ orders all target nodes by decreasing $s_R$, with node ID breaking ties. Under Raw, reviewers start with the least confident predictions and stop after $H$ nodes: $R_H$ is the set they can reach, while the remaining $N-H$ nodes stay unreviewed. \method\ changes which nodes enter this human review window, not its size $H$.

Let $J$ rank all $N$ nodes using the candidate score (Section~\ref{sec:exchange}). Removing nodes already in $R_H$ leaves the eligible pool $\mathcal U\setminus R_H$. For a preselected exchange fraction $\tau\in(0,1)$, the \emph{candidate set} $C$ contains the first $L=\min(N-H,\lfloor\tau H\rfloor)$ eligible nodes in $J$ order. Thus $|C|=L$ and $C\cap R_H=\varnothing$. Only these $L$ candidates enter the paired comparison.

The \emph{reviewed set} $S\subseteq\mathcal U$, $|S|=H$, contains all nodes actually reviewed by \method: paid audit nodes, retained Raw nodes, and the continuation selected by the audit. A corrected prediction earns one unit:
\begin{equation*}
u_i=\ind\{y_i\ne\hat y_i\},\qquad
U(S)=\sum_{i\in S}u_i,\qquad
G=U(S)-U(R_H).
\end{equation*}
We compare complete review policies, whether they follow fixed rankings or adapt to paid labels. The objective is $G>0$: more errors corrected with the same budget. Missed attacks and false alarms have equal reward. Paid audit nodes remain in $S$ even if replacement is declined.

Let $v_i=1$ for a subtype absent from training and validation, and $0$ otherwise. Its contribution to net gain is
\begin{equation*}
G_{\mathrm{OOD}}=\sum_{i\in S}u_iv_i-\sum_{i\in R_H}u_iv_i.
\end{equation*}
This measures corrected OOD errors, not OOD counts or new types.

Known training labels fit the classifier and density; known validation labels select density settings. Scores and configuration are frozen before review, which reveals only requested class labels. Unqueried labels and OOD flags $v_i$ are reserved for offline evaluation. Unlabelled graph covariates may be transductive (Appendix~\ref{app:data}).

\begin{table}[!b]
\centering
\caption{Notation. $R,J$ rank all $N$ nodes; $C$ is the $L$-node candidate set, and $S$ is the $H$-node reviewed set.}\label{tab:symbols}
\begin{tabular}{lp{0.77\textwidth}}\toprule
Symbol & Meaning\\\midrule
$\mathcal U,N,K$ & Target node pool, its size, and number of known review classes.\\
$p_i,p_{ik},\hat y_i,y_i$ & Class probabilities, class-$k$ probability, predicted label and human label.\\
$H,s_R(i),R,R_H$ & Reviewable node limit, Raw uncertainty, full Raw ordering, and its first $H$ nodes.\\
$S,u_i,U(S)$ & Reviewed set ($|S|=H$), correction indicator, and correction count.\\
$v_i,G,G_{\mathrm{OOD}}$ & Unseen-subtype flag; overall and OOD net corrections relative to Raw.\\
$\bar p_i,x_i,z_i$ & Mean neighbor posterior, input features and standardized graph token.\\
$M_k,\gamma_{ib},w_{kb}(x)$ & Number of class-$k$ Gaussian components, responsibility, and node-specific weight.\\
$\pi_{kb},\mu_{kb},\Sigma_{kb}$ & Class-wide weight, mean and covariance of Gaussian component $b$ in class $k$.\\
$g_k(x),\eta_k$ & Feature gate and validation-selected conditioning strength.\\
$f_k(z\mid x),s_J(i)$ & Conditional graph token density and relational surprise score.\\
$q_i,J$ & Priority-adjusted review score and its descending candidate ordering.\\
$\tau,L$ & Exchange fraction and block size $L=\min(N-H,\lfloor\tau H\rfloor)$.\\
$C,D$ & Candidate set outside $R_H$ and proposed removals within $R_H$, each of size $L$.\\
$B$ & Shared Raw set $R_H\setminus D$; retain its first $H-L-m$ nodes after $m$ audited pairs.\\
$m,I_m,\delta_j,p_m$ & Audited pair count and indices; pair reward difference; remaining-gain probability.\\
$A$ & Authorization indicator: $1$ for adopting the remaining exchange, $0$ otherwise.\\\bottomrule
\end{tabular}
\end{table}

\section{The PairAudit Model}
\subsection{Graph tokens and feature-conditioned densities}\label{sec:token}
We construct a graph token from a node’s prediction and the average prediction of its available one-hop neighbors. Let $\bar p_i$ be the mean class-probability vector of node $i$'s available one-hop neighbors, using $p_i$ for an isolated node. Apply the isometric log-ratio transform (ilr) \citep{egozcue2003ilr} to each vector, mapping its $K$ probabilities to $K-1$ nonredundant coordinates:
\begin{equation*}
t_i=[\operatorname{ilr}(p_i),\operatorname{ilr}(\bar p_i)],\qquad
z_i=(t_i-\mu_{\mathrm{train}})/\max(s_{\mathrm{train}},10^{-6}).
\end{equation*}
The graph token $t_i$ has $d=2(K-1)$ coordinates. Training means $\mu_{\mathrm{train}}$ and standard deviations $s_{\mathrm{train}}$ standardize it to $z_i\in\mathbb R^d$; division and the floor are coordinate-wise.

To model familiar patterns in these tokens, we fit a Gaussian mixture within each known class. For known class $k$, let $T_k$ and $V_k$ contain its training and validation nodes. A Gaussian mixture model (GMM) uses $M_k$ components indexed by $b=1,\ldots,M_k$. $M_k$ is the \emph{number of Gaussian components}, not classes or attack subtypes; components approximate different patterns of node and neighborhood predictions within a class.

For each candidate value of $M_k$, hold that count fixed and fit the GMM on $T_k$. Component $b$ has class-wide weight $\pi_{kb}>0$, mean $\mu_{kb}\in\mathbb R^d$ and positive definite covariance $\Sigma_{kb}$, with $\sum_b\pi_{kb}=1$. The component density and training log likelihood to maximize are
\begin{align*}
\phi_d(z;\mu,\Sigma)&=\frac{\exp[-\tfrac12(z-\mu)^\top\Sigma^{-1}(z-\mu)]}{(2\pi)^{d/2}|\Sigma|^{1/2}},
\\
\ell_k&=\sum_{i\in T_k}\log\sum_{b=1}^{M_k}\pi_{kb}\phi_d(z_i;\mu_{kb},\Sigma_{kb}). 
\end{align*}
Full covariance captures dependencies between token coordinates. After K-means initialization, expectation-maximization (EM) alternates two steps. The E step computes responsibility $\gamma_{ib}$, the soft assignment of $z_i$ to component $b$:
\begin{equation*}
\gamma_{ib}=\frac{\pi_{kb}\phi_d(z_i;\mu_{kb},\Sigma_{kb})}
{\sum_{c=1}^{M_k}\pi_{kc}\phi_d(z_i;\mu_{kc},\Sigma_{kc})},\qquad i\in T_k.\label{eq:responsibility}
\end{equation*}
Let $N_b=\sum_{i\in T_k}\gamma_{ib}$ be its effective membership. The M step updates
\begin{align*}
\pi_{kb}^{\mathrm{new}}&=N_b/|T_k|,\qquad \mu_{kb}^{\mathrm{new}}=N_b^{-1}\sum_{i\in T_k}\gamma_{ib}z_i,\nonumber\\
\Sigma_{kb}^{\mathrm{new}}&=N_b^{-1}\sum_{i\in T_k}\gamma_{ib}(z_i-\mu_{kb}^{\mathrm{new}})(z_i-\mu_{kb}^{\mathrm{new}})^\top+\epsilon I,\label{eq:em-update}
\end{align*}
where $I$ is the $d$-dimensional identity and $\epsilon=10^{-4}$ prevents singular covariances. EM yields a local fit. After fitting each admissible count, select the fitted GMM with the highest mean log likelihood on $V_k$. Thus training estimates parameters \emph{for a fixed count}; validation selects the count and its fitted parameters. The selected $M_k,\pi_{kb},\mu_{kb},\Sigma_{kb}$ are then held fixed.

The fitted mixture describes familiar patterns within each class. Its class-wide weights, however, may not suit every node. The weights $\pi_{kb}$ describe how common each pattern is in class $k$, but may not suit every node. For example, disagreement with neighbors may be rare among benign nodes overall yet familiar for benign shared servers. A component describing this server pattern can have a small class-wide weight, making an ordinary server look surprising.

We use a feature gate $g_k(x)=(g_{k1}(x),\ldots,g_{kM_k}(x))$ to adapt the weights to a node's input features $x$. Its outputs are nonnegative and sum to one. For server-like features, it can assign more weight to the familiar server pattern. This changes the \emph{weighted sum} of the fixed component densities, not the Gaussian means or covariances. The token must still fit that pattern to receive higher density.

Learn $g_k$ from known training pairs $(x_i,z_i)$ by maximizing conditional log likelihood:
\begin{equation}
\max_{g_k}\sum_{i\in T_k}\log\sum_{b=1}^{M_k}
\frac{g_{kb}(x_i)+\lambda\pi_{kb}}{1+\lambda}\phi_d(z_i;\mu_{kb},\Sigma_{kb}),\quad\lambda=10^{-4}.\label{eq:gate-loss}
\end{equation}
This objective rewards higher density for relationships observed with those features. A feature-tree gate is initialized from $\gamma_{ib}$, then optimized for this likelihood (Appendix~\ref{app:density}).

Blend the learned weights with the class-wide weights:
\begin{equation*}
w_{kb}(x)=(1-\eta_k)\pi_{kb}+\eta_k\frac{g_{kb}(x)+\lambda\pi_{kb}}{1+\lambda}.\label{eq:gate-normalization}
\end{equation*}
Here $\lambda$ keeps weights positive, and $\eta_k$ controls how strongly features adjust them. Although $\eta_k$ is shared within class $k$, $w_{kb}(x)$ varies by node. The conditional density is
\begin{equation*}
f_k(z\mid x)=\sum_{b=1}^{M_k}w_{kb}(x)\phi_d(z;\mu_{kb},\Sigma_{kb}).
\end{equation*}
Select the gate checkpoint and $\eta_k\in\{0,.25,.5,.75,1\}$ by mean conditional log likelihood on $V_k$. At $\eta_k=0$, the class-wide mixture is recovered; with one component, weights cannot vary. Neither stage uses OOD labels or simulated audit gains.

The gate uses only $x$; each Gaussian still checks the actual relationship $z$. Familiar features cannot excuse a mismatching token. For fixed $x$, $\int f_k(z\mid x)\,dz=\sum_bw_{kb}(x)=1$. Training rewards aggregate likelihood, not necessarily higher density for every node; high density does not certify a correct prediction.

We rank target nodes for review using the fitted conditional density, evaluating each token under its predicted class density since the true class is unknown. For a target node, use its predicted class and define surprise $a_i=-\log f_{\hat y_i}(z_i\mid x_i)$. Higher surprise means lower density. Known validation nodes $v\in V_k$ provide reference values $a_v=-\log f_k(z_v\mid x_v)$. The ranking score is
\begin{equation*}
s_J(i)=\frac{\sum_{v\in V_k}\ind\{a_v\le a_i\}+\tfrac12+
\tfrac{1}{2\pi}\arctan((a_i-m_k)/\sigma_k)}{|V_k|+1},\quad k=\hat y_i,\label{eq:score}
\end{equation*}
where $m_k$ and $\sigma_k$ are the reference median and standard deviation, with $\sigma_k$ floored at $10^{-8}$. The reference aligns class scales. When the validation comparison gives two nodes the same score, the small arctan term ranks the node with higher surprise first. Larger $s_J$ means greater relational surprise. Appendix~\ref{app:density} specifies numerical tie handling.

\subsection{Ordinary labels and sequential continuation}\label{sec:audit}
From the fixed candidate ranking $J$, take $C$ as the first $L$ nodes outside $R_H$, and $D$ as the $L$ nodes inside $R_H$ ranked lowest by $J$. The remainder $B=R_H\setminus D$ follows Raw order; these disjoint sets are fixed before querying labels. Appendix~\ref{sec:exchange} defines priorities for different applications and evaluates giving priority to nodes predicted as attacks when false alarms are costly.

\begin{figure}[t]
\centering\includegraphics[width=\textwidth]{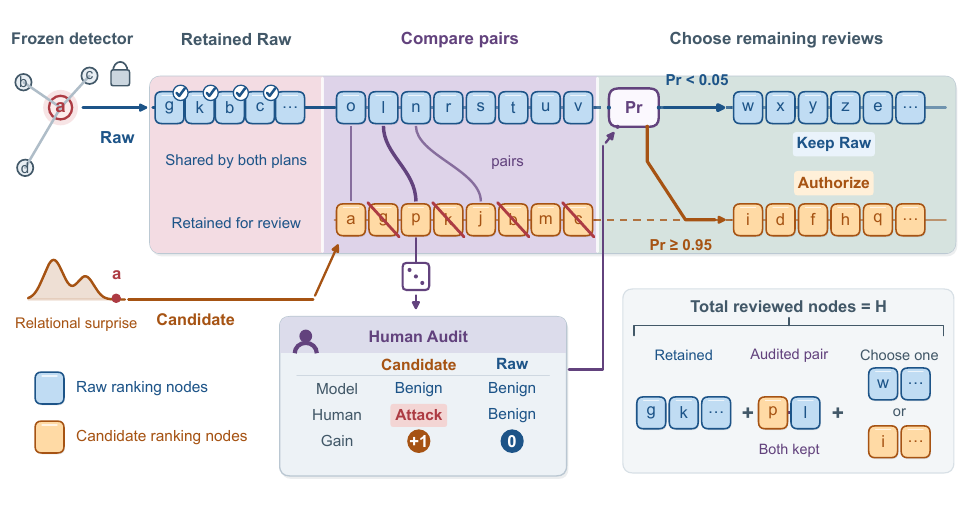}
\caption{Raw (upper row) and candidate replacements (lower row). Check marks indicate retention; crossed nodes overlap Raw's first $H$. The die denotes uniform pair sampling; p/l gives candidate gain $+1$ and Raw gain $0$. Feedback updates $p_m$ (Equation~(\ref{eq:predictive-tail})); authorization also requires three non-ties. Both paid endpoints count within $H$ (inset; Equation~(\ref{eq:completion})). Dashed lines connect queue segments. Node counts are illustrative.}\label{fig:workflow}
\end{figure}

Sort $C$ and $D$ separately by increasing predicted class ID, increasing $s_R(i)$, then node ID. Match corresponding positions to obtain $(c_j,d_j)$, $j=1,\ldots,L$, with $c_j\in C$ and $d_j\in D$. This fixes pair membership, not review order. Draw one uniform random permutation of the pairs and review it without replacement. Each pair costs two ordinary class labels and gives
\begin{equation*}
\delta_j=u_{c_j}-u_{d_j}\in\{-1,0,1\}.
\end{equation*}
A $+1$ means the frozen detector misclassified only $c_j$; $-1$ means only $d_j$; $0$ means both or neither. Pairing never uses human labels. Appendix~\ref{app:pairing} tests alternative pairings.

After $m$ pairs, $2m$ distinct nodes have been reviewed. Let $n=(n_+,n_-,n_0)$ count the observed signs, with $n_++n_-+n_0=m$. The working model assigns unknown probabilities $\theta=(\theta_+,\theta_-,\theta_0)$ to a candidate win, Raw win and tie. A Dirichlet$(1/2,1/2,1/2)$ Jeffreys prior gives
\begin{equation*}
\theta\mid n\sim\operatorname{Dirichlet}(\alpha_+,\alpha_-,\alpha_0),
\qquad \alpha_t=n_t+\tfrac12,\quad t\in\{+,-,0\}.
\end{equation*}

For $r=L-m$ unreviewed pairs, let $X_+,X_-,X_0$ count their unknown signs. Choosing candidate rather than Raw endpoints adds $X_+-X_-$ corrections, giving predictive probability $p_m=\Pr(X_+>X_-\mid n)$. Set $V=X_++X_-$, the remaining non-tie count. Writing $\operatorname{BB}(r,a,b)$ for a beta-binomial count with $r$ trials and beta parameters $a,b$,
\begin{equation}
\begin{aligned}
V\mid n&\sim\operatorname{BB}(r,\alpha_++\alpha_-,\alpha_0),\\
X_+\mid V=v,n&\sim\operatorname{BB}(v,\alpha_+,\alpha_-).
\end{aligned}\label{eq:predictive}
\end{equation}
For $V=v$, positive gain requires $X_+>v/2$. Averaging over possible non-tie counts gives
\begin{equation}
p_m=\sum_{v=0}^r\Pr(V=v\mid n)\left[1-F_{\operatorname{BB}(v,\alpha_+,\alpha_-)}(\lfloor v/2\rfloor)\right],\label{eq:predictive-tail}
\end{equation}
where $F$ is the indicated distribution's cumulative probability. The $v=0$ term is zero: all ties give no gain.

We use \(p_m\) to decide whether to adopt the remaining exchange. Decisions occur after label batches, and each is called a look. Nominal \emph{cumulative} costs are $5,10,20,30,50\%$ of $H$; actual cost is $2m/H$ (rounding and caps: Appendix~\ref{app:algorithm}). Authorize if $p_m\ge.95$ and $n_++n_-\ge3$, requiring three observed non-ties. Reject if $p_m<.05$. Otherwise continue, rejecting if the last look remains unresolved. This working-model evidence concerns the remaining exchange, not total workflow gain.

Stopping ends pair sampling, not the remaining reviews. Let $I_m$ contain reviewed pair indices and $\overline I_m=\{1,\ldots,L\}\setminus I_m$. Write $C_{I_m}=\{c_j:j\in I_m\}$, analogously for $D$ and unreviewed indices. Let $B_{1:h}$ be the first $h$ nodes of $B$ in Raw order (empty for $h=0$). Keep all paid endpoints and form
\begin{equation}
S=C_{I_m}\cup D_{I_m}\cup
\begin{cases}C_{\overline I_m},&\text{authorized},\\D_{\overline I_m},&\text{otherwise},\end{cases}
\ \cup B_{1:H-L-m}.\label{eq:completion}
\end{equation}
Thus $2m$ audited nodes, $L-m$ continuation nodes and $H-L-m$ retained nodes sum to $H$. Keeping both pair endpoints costs one extra slot per pair, displacing $m$ nodes from the tail of $B$. Feasible looks require $m<L$ and $m\le H-L$; if $L<2$, use $R_H$ without audit queries.

\section{What the Audit Establishes}
\begin{proposition}[Budget and gain identity]\label{prop:budget}
Equation~(\ref{eq:completion}) gives exactly $H$ unique reviews at each feasible look. Let $A\in\{0,1\}$ indicate authorization and $B_{\mathrm{lost}}=B\setminus B_{1:H-L-m}$. Then
\begin{equation*}
G=U(C_{I_m})-U(B_{\mathrm{lost}})+A\sum_{j\notin I_m}\delta_j.
\end{equation*}
\end{proposition}
The three disjoint blocks total $H$; subtracting $U(B)+U(D)$ gives the identity. It separates corrections found during audit, corrections displaced by its cost, and the remaining exchange gain. No node independence is required.

A favorable exchange may not cover audit costs. We distinguish harmful authorization, which yields fewer subsequent corrections than Raw after the same paid queries, from total workflow loss, $G<0$. Appendix~\ref{app:proofs} defines the comparison and derives the posterior calculation.

We next consider the sampling distribution for the fixed pairs. For frozen pairs with sign totals $K=(K_+,K_-,K_0)$, uniform sampling of $m$ pairs gives
\begin{equation*}
\Pr(n\mid K)=\frac{\prod_{t\in\{+,-,0\}}\binom{K_t}{n_t}}{\binom{L}{m}},\qquad\sum_t n_t=m.\label{eq:sampling-law}
\end{equation*}
This law holds even with clustered graph errors \citep{cochran1977sampling}. It does not calibrate the Dirichlet working posterior, which supplies model-based evidence rather than a distribution-free guarantee. 

\section{Experiments}
\subsection{Protocol and questions}
We evaluate correction gains, relational evidence and harmful candidate rejection on four primary security tasks, prevalence and multiclass extensions, and a Physics control. All use $H=\lceil.1N\rceil$ and default $\tau=1/3$. This preset can be adjusted for different datasets, but we do not use each dataset's best tested value for the main results. PairAudit still has room for gains exceeding those presented in the main experiments, as shown in Appendix~\ref{app:tau}. A fit is a frozen backbone and source configuration qualified on known training and validation labels. We average 600 audit randomizations per workload, then weight fits equally; these are not independent attack events. Review reveals ordinary dataset labels at equal cost without added noise; subtype flags are evaluator-only. Tasks were selected retrospectively.

Appendices~\ref{app:data} and~\ref{app:instances} provide data, backbones and seeds. Component count, exchange size and pairing tests appear in Appendices~\ref{sec:component-sensitivity}, \ref{app:tau} and~\ref{app:pairing}; Figure~\ref{fig:tradeoff} shows the exchange size benefit--cost tradeoff.

\subsection{Which errors enter the paid review budget?}
\begin{table}[!htbp]
\centering

\setlength{\tabcolsep}{4pt}
\caption{Equal-budget corrections and net gains over Raw. $G_{\mathrm{OOD}}$ counts only unseen-subtype errors.}\label{tab:main}
\begin{tabular}{lrrrrrr}\toprule
Scenario & Fits & Raw & PairAudit & $G$ & $G_{\mathrm{OOD}}$ & Auth. (\%)\\\midrule
UNSW / Fuzzers & 20 & 224.95 & 327.90 & +102.95 & +105.72 & 100.000\\
CICIDS / Hulk & 10 & 1097.60 & 1887.39 & +789.79 & +830.80 & 100.000\\
X-IIoTID / RDOS & 3 & 440.33 & 780.63 & +340.30 & +461.00 & 100.000\\
TON / XSS & 3 & 383.33 & 463.35 & +80.01 & +102.02 & 100.000\\
\bottomrule\end{tabular}
\par\vspace{3pt}\begin{minipage}{\textwidth} All paid reviews are included. Auth. is the percentage of runs authorizing the candidate.\end{minipage}
\end{table}

\begin{table}[!htbp]
\centering

\setlength{\tabcolsep}{4pt}
\caption{Target population and review budget used before deciding.}\label{tab:budget}
\begin{tabular}{lrrrrrr}\toprule
Scenario & $N$ & $H$ & $H/N$ (\%) & Labels & $2m/H$ (\%) & On auth. (\%)\\\midrule
UNSW / Fuzzers & 25,305 & 2,531 & 10.00 & 144.4 & 5.71 & 5.71\\
CICIDS / Hulk & 27,114 & 2,712 & 10.00 & 136.0 & 5.01 & 5.01\\
X-IIoTID / RDOS & 14,187 & 1,419 & 10.00 & 72.0 & 5.07 & 5.07\\
TON / XSS & 11,407 & 1,141 & 10.00 & 78.9 & 6.92 & 6.92\\
\bottomrule\end{tabular}
\par\vspace{3pt}\begin{minipage}{\textwidth} Labels and $2m/H$ average all runs; the last column averages authorized runs only. Final review cost is always $H$.\end{minipage}
\end{table}

All four primary scenarios show positive mean $G$ and $G_{\mathrm{OOD}}$ after all pair costs, and both fit-level means are positive in all 36 configurations. UNSW adds 102.95 net corrections to Raw's 224.95, while CICIDS adds 789.79 to 1,097.60. The effect is meaningful because these are net errors Raw would not recover within the same capacity. In X-IIoTID / RDOS, $G_{\mathrm{OOD}}=461.00$ exceeds $G=340.30$: 120.70 known-object corrections are lost, but the unseen-type benefit more than compensates. This is a tradeoff, not a count of additional unknown objects.

\begin{figure}[t]
\centering\includegraphics[width=\textwidth]{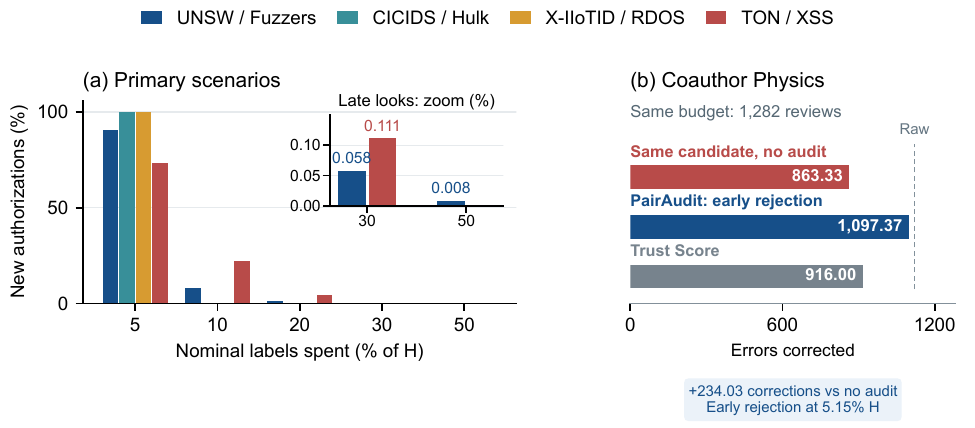}
\caption{(a) New authorizations at each nominal look, as percentages of all runs per task; the inset magnifies rare late decisions. (b) Early rejection on Coauthor Physics, with $H=1{,}282$ reviews. Adopting the same candidate exchange $B\cup C$ without audit corrects 863.33 errors; PairAudit corrects 1,097.37, including audit costs. Raw corrects 1,120.33. Trust Score is a separate ranking reference.}\label{fig:cost}
\end{figure}

Decisions generally leave most of the budget available for continuation (Table~\ref{tab:budget}). CICIDS and X-IIoTID / RDOS authorize at the first feasible look in all runs; TON requires 6.92\% of $H$ on average. All 21,600 primary runs authorize. Authorization is a fraction of complete audit runs, not a fraction of labels judged beneficial. Figure~\ref{fig:cost} shows newly occurring decisions by look. The primary tasks show neither harmful authorized continuation nor negative total gain in the evaluated runs.

On Physics, PairAudit rejects its candidate at the first look in all 1,800 runs after 66 labels ($5.15\%$ of $H$). Compared with adopting the same exchange without audit, it preserves 234.03 corrections at the same budget (Figure~\ref{fig:cost}b). Its mean loss relative to Raw is 22.97, versus 204.33 for Trust Score; the latter is a separate scorer comparison.

\subsection{Comparison with independent review rankings}
\begin{table}[!htbp]
\centering

\setlength{\tabcolsep}{4pt}
\caption{Independent review strategies at equal budget. Entries are $G/G_{\mathrm{OOD}}$.}\label{tab:external}
\begin{tabular}{lrrrr}\toprule
Method & UNSW / Fuzzers & CICIDS / Hulk & X-IIoTID / RDOS & TON / XSS\\\midrule
Raw & 0.00 / 0.00 & 0.00 / 0.00 & 0.00 / 0.00 & 0.00 / 0.00\\
Trust Score & -23.90 / -29.90 & +886.60 / +817.20 & -41.00 / +0.67 & +166.67 / +135.67\\
Energy & -37.15 / -36.05 & -270.40 / -263.80 & -12.67 / +0.00 & +26.33 / +29.33\\
GNNSafe & -61.40 / -50.60 & -546.60 / -549.70 & -98.67 / +0.00 & -13.33 / +26.67\\
GRASP & -11.20 / +3.90 & +224.10 / +425.70 & -155.67 / +0.00 & -138.33 / -87.00\\
PairAudit & \textbf{+102.95 / +105.72} & \textbf{+789.79 / +830.80} & \textbf{+340.30 / +461.00} & \textbf{+80.01 / +102.02}\\
\bottomrule\end{tabular}
\par\vspace{3pt}\begin{minipage}{\textwidth} External methods use their own first $H$ nodes, without our audit. All methods in the CICIDS column share ten checkpoint reruns; Appendix~\ref{app:baselines} gives details.\end{minipage}
\end{table}

Table~\ref{tab:external} compares complete policies, not isolated scores. Raw retention and paid audits are parts of PAIRAUDIT's policy. Trust Score is strongest in CICIDS and TON on total corrections, so \method\ is not uniformly superior. On CICIDS, however, \method\ corrects 13.60 more unseen-attack errors while giving up 96.81 total corrections. On X-IIoTID / RDOS, Trust Score loses 41.00 total corrections and adds only 0.67 OOD corrections, whereas \method\ improves both. These differences show why overall and unseen-attack corrections should be assessed together. Appendix~\ref{app:baselines} gives public settings and input limitations.

\subsection{What does relational conditioning contribute?}
Appendix~\ref{app:ablations} reports the full controls in Table~\ref{tab:mechanism-full}. Feature-conditioned weights improve mean gain over fixed weights on all four primary tasks. TON provides the clearest evidence for relational information: the graph token model yields $G=+80.01$, versus $-39.86$ using only node coordinates and $+65.36$ after matched neighborhood shuffling. The benefit is task dependent: removing neighborhood coordinates improves X-IIoTID / RDOS, and matched shuffling does not reduce gain on CICIDS.

\subsection{Prevalence stress tests and multiclass review}
\begin{table}[!htbp]
\centering

\setlength{\tabcolsep}{2.5pt}
\caption{Stress tests and multiclass review, relative to the equal-budget Raw baseline for each pool.}\label{tab:extension}
\begin{tabular}{lrrrrrrrr}\toprule
Task / pool & $N$ & $H$ & OOD (\%) & $\rho_{\mathrm{err}}$ (\%) & $G$ & $G_{\mathrm{OOD}}$ & Auth. (\%) & Cost (\%)\\\midrule
IoT23 / ARP (1\%) & 13,693 & 1,370 & 1.00 & 0.514 & +58.59 & +20.03 & 99.96 & 9.83\\
IoMT / ARP (full) & 6,554 & 656 & 20.90 & 12.084 & +44.19 & +44.51 & 100.00 & 8.61\\
IoMT / ARP (95\%) & 1,442 & 145 & 95.01 & 54.924 & +44.33 & +44.33 & 100.00 & 5.53\\
BoT / DDoS-TCP & 7,482 & 749 & 66.65 & 65.143 & +37.81 & +101.60 & 99.50 & 13.05\\
\bottomrule\end{tabular}
\par\vspace{3pt}\begin{minipage}{\textwidth} $\rho_{\mathrm{err}}$: OOD errors divided by $N$; cost: $2m/H$. ARP tasks are binary; BoT uses three classes without priority. Pool construction and configurations are in Appendices~\ref{app:data} and~\ref{app:instances}.\end{minipage}
\end{table}

CIC IoT 2023 / MITM-ArpSpoofing yields $58.59$ net corrections, including $20.03$ on unseen attacks, with only $1\%$ OOD objects: benefit need not require high OOD prevalence. IoMT / ARP also improves both its full and $95\%$ OOD pools, mainly through unseen-attack corrections (Table~\ref{tab:extension}). This comparison is descriptive because datasets and graph contexts differ. In multiclass BoT-IoT, PairAudit increases corrections from $407.00$ to $444.81$ with $99.50\%$ authorization while retaining the DDoS/DoS/reconnaissance review task. Its $101.60$ additional OOD corrections outweigh $63.79$ fewer known-subtype corrections.  Appendices~\ref{app:data} and~\ref{app:instances} give construction and run-level losses.

\FloatBarrier
\section{Limitations and Conclusion}
\method\ turns graph-based evidence into actionable review decisions under distribution shift. It finds potentially overlooked errors and uses ordinary class labels to decide whether a candidate queue deserves the remaining review budget, without retraining the detector or changing the review task. Across the four primary security tasks, it delivers positive mean overall and unseen-attack correction gains with high authorization rates and decisions that use only a small share of the budget. The rejection control further shows how early feedback can limit losses from an unfavorable proposal. Together, these results support combining Raw retention, relational evidence and paid feedback within a single review policy.

The benefit of graph tokens varies across tasks, and our current formulation assumes equal review costs and gives each corrected error the same value. The evaluation also does not explicitly model reviewer mistakes. Extending the framework to different review times, error costs, and uncertain labels would require changes to the review objective and audit rule. These limitations motivate further work on cost sensitive review, noisy human feedback, and stronger statistical guarantees.

\section*{AI Use Statement}
AI tools assisted with manuscript preparation. Reported quantities are computed from saved predictions and review ledgers. AI outputs are not experimental observations or sources of authority.
\section*{Reproducibility Statement}
The supplement specifies the Gaussian and gate estimation, audit computation, data roles and all configurations. Code implements the frozen scoring and paid review procedures; evidence files record frozen inputs, audit seeds, decisions, review-set checksums and aggregate metrics. Appendix~\ref{app:repro} describes verification and source provenance. No paid human-subject study is claimed.
\bibliography{references}

\begin{thebibliography}{25}
\providecommand{\natexlab}[1]{#1}
\providecommand{\url}[1]{\texttt{#1}}
\expandafter\ifx\csname urlstyle\endcsname\relax
  \providecommand{\doi}[1]{doi: #1}\else
  \providecommand{\doi}{doi: \begingroup \urlstyle{rm}\Url}\fi

\bibitem[Al-Hawawreh et~al.(2022)Al-Hawawreh, Sitnikova, and Aboutorab]{alhawawreh2022xiiotid}
Muna Al-Hawawreh, Elena Sitnikova, and Neda Aboutorab.
\newblock {X-IIoTID}: A connectivity-agnostic and device-agnostic intrusion data set for industrial internet of things.
\newblock \emph{IEEE Internet of Things Journal}, 9\penalty0 (5):\penalty0 3962--3977, 2022.
\newblock \doi{10.1109/JIOT.2021.3102056}.

\bibitem[Bai et~al.(2025)Bai, Du, Rainey, Parameswaran, and Li]{bai2025human}
Haoyue Bai, Xuefeng Du, Katie Rainey, Shibin Parameswaran, and Yixuan Li.
\newblock Out-of-distribution learning with human feedback.
\newblock \emph{Transactions on Machine Learning Research}, 2025.

\bibitem[Bishop(1994)]{bishop1994mixture}
Christopher~M. Bishop.
\newblock Mixture density networks.
\newblock Technical Report NCRG/94/004, Aston University, 1994.

\bibitem[{Canadian Institute for Cybersecurity}(2023)]{ciciot2023}
{Canadian Institute for Cybersecurity}.
\newblock {CIC IoT Dataset 2023}.
\newblock University of New Brunswick, 2023.
\newblock URL \url{https://www.unb.ca/cic/datasets/iotdataset-2023.html}.

\bibitem[{Canadian Institute for Cybersecurity}(2024)]{ciciomt2024}
{Canadian Institute for Cybersecurity}.
\newblock {CIC IoMT Dataset 2024}.
\newblock University of New Brunswick, 2024.
\newblock URL \url{https://www.unb.ca/cic/datasets/iomt-dataset-2024.html}.

\bibitem[Cochran(1977)]{cochran1977sampling}
William~G. Cochran.
\newblock \emph{Sampling Techniques}.
\newblock John Wiley \& Sons, 3 edition, 1977.

\bibitem[Egozcue et~al.(2003)Egozcue, Pawlowsky-Glahn, Mateu-Figueras, and Barcel{\'o}-Vidal]{egozcue2003ilr}
Juan~Jos{\'e} Egozcue, Vera Pawlowsky-Glahn, Gl{\`o}ria Mateu-Figueras, and Carles Barcel{\'o}-Vidal.
\newblock Isometric logratio transformations for compositional data analysis.
\newblock \emph{Mathematical Geology}, 35\penalty0 (3):\penalty0 279--300, 2003.
\newblock \doi{10.1023/A:1023818214614}.

\bibitem[Errica et~al.(2021)Errica, Bacciu, and Micheli]{errica2021graph}
Federico Errica, Davide Bacciu, and Alessio Micheli.
\newblock Graph mixture density networks.
\newblock In \emph{International Conference on Machine Learning}, pp.\  3025--3035, 2021.

\bibitem[Hendrycks \& Gimpel(2017)Hendrycks and Gimpel]{hendrycks2017baseline}
Dan Hendrycks and Kevin Gimpel.
\newblock A baseline for detecting misclassified and out-of-distribution examples in neural networks.
\newblock In \emph{International Conference on Learning Representations}, 2017.

\bibitem[Jiang et~al.(2018)Jiang, Kim, Guan, and Gupta]{jiang2018trust}
Heinrich Jiang, Been Kim, Melody Guan, and Maya Gupta.
\newblock To trust or not to trust a classifier.
\newblock In \emph{Advances in Neural Information Processing Systems}, 2018.

\bibitem[Koroniotis et~al.(2019)Koroniotis, Moustafa, Sitnikova, and Turnbull]{koroniotis2019bot}
Nickolaos Koroniotis, Nour Moustafa, Elena Sitnikova, and Benjamin Turnbull.
\newblock Towards the development of realistic botnet dataset in the internet of things for network forensic analytics: Bot-iot dataset.
\newblock \emph{Future Generation Computer Systems}, 100:\penalty0 779--796, 2019.

\bibitem[Kossen et~al.(2021)Kossen, Farquhar, Gal, and Rainforth]{kossen2021active}
Jannik Kossen, Sebastian Farquhar, Yarin Gal, and Tom Rainforth.
\newblock Active testing: Sample-efficient model evaluation.
\newblock In \emph{International Conference on Machine Learning}, pp.\  5753--5763, 2021.

\bibitem[Laroche et~al.(2019)Laroche, Trichelair, and Des~Combes]{laroche2019spibb}
Romain Laroche, Paul Trichelair, and Remi~Tachet Des~Combes.
\newblock Safe policy improvement with baseline bootstrapping.
\newblock In \emph{International Conference on Machine Learning}, pp.\  3652--3661, 2019.

\bibitem[Liu et~al.(2020)Liu, Wang, Owens, and Li]{liu2020energy}
Weitang Liu, Xiaoyun Wang, John~D. Owens, and Yixuan Li.
\newblock Energy-based out-of-distribution detection.
\newblock In \emph{Advances in Neural Information Processing Systems}, 2020.

\bibitem[Ma et~al.(2024)Ma, Sun, Ding, Liu, and Wu]{ma2024revisiting}
Longfei Ma, Yiyou Sun, Kaize Ding, Zemin Liu, and Fei Wu.
\newblock Revisiting score propagation in graph out-of-distribution detection.
\newblock In \emph{Advances in Neural Information Processing Systems}, 2024.

\bibitem[Moustafa(2021)]{moustafa2021ton}
Nour Moustafa.
\newblock A new distributed architecture for evaluating {AI}-based security systems at the edge: Network {TON\_IoT} datasets.
\newblock \emph{Sustainable Cities and Society}, 72:\penalty0 102994, 2021.
\newblock \doi{10.1016/j.scs.2021.102994}.

\bibitem[Moustafa \& Slay(2015)Moustafa and Slay]{moustafa2015unsw}
Nour Moustafa and Jill Slay.
\newblock {UNSW-NB15}: A comprehensive data set for network intrusion detection systems.
\newblock In \emph{Military Communications and Information Systems Conference}, pp.\  1--6, 2015.
\newblock \doi{10.1109/MilCIS.2015.7348942}.

\bibitem[Roy et~al.(2024)Roy, Shu, Li, Yang, Elshocht, Smeets, and Li]{roy2024gadnr}
Amit Roy, Juan Shu, Jia Li, Carl Yang, Olivier Elshocht, Jeroen Smeets, and Pan Li.
\newblock {GAD-NR}: Graph anomaly detection via neighborhood reconstruction.
\newblock In \emph{ACM International Conference on Web Search and Data Mining}, pp.\  576--585, 2024.
\newblock \doi{10.1145/3616855.3635767}.

\bibitem[Sharafaldin et~al.(2018)Sharafaldin, Lashkari, and Ghorbani]{sharafaldin2018toward}
Iman Sharafaldin, Arash~Habibi Lashkari, and Ali~A. Ghorbani.
\newblock Toward generating a new intrusion detection dataset and intrusion traffic characterization.
\newblock In \emph{International Conference on Information Systems Security and Privacy}, pp.\  108--116, 2018.
\newblock \doi{10.5220/0006639801080116}.

\bibitem[Shchur et~al.(2018)Shchur, Mumme, Bojchevski, and G{\"u}nnemann]{shchur2018pitfalls}
Oleksandr Shchur, Maximilian Mumme, Aleksandar Bojchevski, and Stephan G{\"u}nnemann.
\newblock Pitfalls of graph neural network evaluation.
\newblock \emph{arXiv preprint arXiv:1811.05868}, 2018.

\bibitem[Song et~al.(2007)Song, Wu, Jermaine, and Ranka]{song2007conditional}
Xiuyao Song, Mingxi Wu, Christopher Jermaine, and Sanjay Ranka.
\newblock Conditional anomaly detection.
\newblock \emph{IEEE Transactions on Knowledge and Data Engineering}, 19\penalty0 (5):\penalty0 631--645, 2007.
\newblock \doi{10.1109/TKDE.2007.1009}.

\bibitem[Stouffer et~al.(2023)Stouffer, Pease, Tang, Zimmerman, Pillitteri, Lightman, Hahn, Saravia, Sherule, and Thompson]{stouffer2023ot}
Keith Stouffer, Michael Pease, CheeYee Tang, Timothy Zimmerman, Victoria Pillitteri, Suzanne Lightman, Adam Hahn, Stephanie Saravia, Aslam Sherule, and Michael Thompson.
\newblock Guide to operational technology ({OT}) security.
\newblock Technical Report SP 800-82 Rev. 3, National Institute of Standards and Technology, 2023.

\bibitem[Vishwakarma et~al.(2024)Vishwakarma, Lin, and Korlakai~Vinayak]{vishwakarma2024taming}
Harit Vishwakarma, Heguang Lin, and Ramya Korlakai~Vinayak.
\newblock Taming false positives in out-of-distribution detection with human feedback.
\newblock In \emph{International Conference on Artificial Intelligence and Statistics}, volume 238 of \emph{Proceedings of Machine Learning Research}, pp.\  1486--1494, 2024.

\bibitem[{Washington University in St. Louis}(2021)]{wustl2021dataset}
{Washington University in St. Louis}.
\newblock {WUSTL-IIoT-2021} dataset for {IIoT} cybersecurity research, 2021.
\newblock URL \url{https://www.cs.wustl.edu/~jain/iiot2/index.html}.

\bibitem[Wu et~al.(2023)Wu, Chen, Yang, and Yan]{wu2023gnnsafe}
Qitian Wu, Yiting Chen, Chenxiao Yang, and Junchi Yan.
\newblock Energy-based out-of-distribution detection for graph neural networks.
\newblock In \emph{International Conference on Learning Representations}, 2023.

\end{thebibliography}
\bibliographystyle{iclr2027_conference}
\clearpage
\appendix
\section{Gaussian Mixtures and Feature-Conditioned Weights}\label{app:density}
\subsection{Coordinate conventions}
Clip probabilities below $10^{-8}$ and renormalize. A $(K-1)\times K$ Helmert matrix $A$ with $AA^\top=I$ and $A\mathbf1=0$ implements $\operatorname{ilr}(p)=A\log p$.

\subsection{EM estimation and component selection}
The GMM implementation uses scikit-learn GaussianMixture: full covariance, two initializations, at most 300 iterations, tolerance $10^{-3}$ and seed 9126101.

Cap fitting at 8,000 training nodes per class, subsampling uniformly with seed $9126100+k$. Declare the candidate counts before target evaluation. A count $M$ requires $M\max(10,2d)$ fit samples, EM convergence, finite validation likelihood and expected membership of at least two per component. Validation ties within $10^{-12}$ favor smaller $M$; no admissible count means fitting fails.

\subsection{Conditional-likelihood gate estimation}
Initialize an ExtraTreesRegressor from training responsibilities: 128 trees, minimum leaf size 32, feature fraction $.7$, seed 9127201. Freeze its partitions. For leaf $\ell$ among $T=128$ trees, set $A_{i\ell}=1/T$ if node $i$ reaches it, else zero. With leaf probabilities $\vartheta_\ell$ summing to one, $g_{kb}(x_i)=\sum_\ell A_{i\ell}\vartheta_{\ell b}$.

Write $\phi_{ib}=\phi_d(z_i;\mu_{kb},\Sigma_{kb})$ and $F_i=\sum_b(g_{kb}(x_i)+\lambda\pi_{kb})\phi_{ib}/(1+\lambda)$. The E step assigns trainable responsibilities
\[
\xi_{i\ell b}=\frac{A_{i\ell}\vartheta_{\ell b}\phi_{ib}}{(1+\lambda)F_i},\qquad
\vartheta_{\ell b}^{\mathrm{new}}=\frac{\sum_i\xi_{i\ell b}}{\sum_{i,c}\xi_{i\ell c}}.
\]
The fixed $\lambda\pi_k$ term receives no trainable responsibility; zero-membership leaves keep their weights. In exact arithmetic, these EM updates do not decrease Equation~(\ref{eq:gate-loss}), which is concave in the leaf weights for fixed partitions and Gaussian components.

Initialize leaf means from $\gamma_i$, floor at $10^{-10}$ and renormalize. Select among updates $\{0,1,2,4,8,16,32,64\}$ and the main-text $\eta_k$ grid by known-validation likelihood. Update zero is initialization; ties within $10^{-12}$ favor the earlier update, then smaller $\eta_k$.

\subsection{Numerical scoring}

Use log-sum-exp, fixed per-row Cholesky arithmetic, eight-decimal log-density rounding and single-threaded gate prediction to keep scores stable across batch sizes.

\FloatBarrier

\section{Execution, Label Access and Numerical Audit}\label{app:algorithm}
\subsection{Candidate ranking}\label{sec:exchange}
Let $\operatorname{rank}_{s_J}(i)$ order nodes by decreasing $s_J$, starting at 1, with node ID breaking ties. For a priority class $c_{\mathrm{priority}}$ chosen before review, define
\begin{equation*}
q_i=2\ind\{\hat y_i=c_{\mathrm{priority}}\}+1-\frac{\operatorname{rank}_{s_J}(i)-1}{N+1}.
\end{equation*}
The rank term lies in $(0,1]$, so adding 2 puts that predicted class first while preserving its score order. Sorting $q_i$ from highest to lowest gives $J$; without a priority class, omit the indicator. Construct $C,D,B$ as in Section~\ref{sec:audit}. This changes candidate and replacement selection, not the uncertainty-based Raw reference. Each corrected error still earns one unit.

The UNSW, CICIDS, X-IIoTID and TON experiments represent settings where missed attacks are especially costly. We therefore set $c_{\mathrm{priority}}=\mathrm{benign}$ to seek attacks hidden among benign predictions. In industrial applications, falsely flagging normal activity can instead prompt costly blocking or isolation; automatic responses must account for their effect on normal operation \citep{stouffer2023ot}. When avoiding these disruptions is the review priority, set $c_{\mathrm{priority}}=\mathrm{attack}$, with the same density model, budget and audit rule. These cost preferences describe intended applications; the datasets do not measure an economic cost ratio.

Table~\ref{tab:attack-priority} evaluates this choice on X-IIoTID / CoAP and WUSTL-IIoT-2021 / UDP. The held-out normal service or protocol is absent from training and validation but retains its benign label, so OOD corrections here recover false alarms on unfamiliar normal activity. Three fits per task each average 600 audit runs; Appendix~\ref{app:industrial} gives construction and selection details.

\begin{table}[!htbp]
\centering
\setlength{\tabcolsep}{4pt}
\caption{Predicted-attack priority on unseen normal activity, relative to equal-budget Raw.}\label{tab:attack-priority}
\begin{tabular}{lrrrrrr}\toprule
Task & $H$ & OOD (\%) & $G$ & $G_{\mathrm{OOD}}$ & Auth. (\%) & Cost (\%)\\\midrule
X-IIoTID / CoAP & 1,419 & 35.24 & +282.47 & +447.33 & 100.00 & 5.07\\
WUSTL-IIoT-2021 / UDP & 710 & 70.51 & +118.64 & +124.03 & 98.39 & 9.12\\
\bottomrule\end{tabular}
\par\vspace{3pt}\begin{minipage}{\textwidth} $G$: net change in all corrected errors; $G_{\mathrm{OOD}}$: the same change restricted to held-out normal activity. Both compare final paid reviews with Raw's first $H$ nodes. Cost is $2m/H$; all decision labels count toward $H$.\end{minipage}
\end{table}

Both gains are positive in every fit mean,  CoAP notably has no negative gain run or harmful authorization in 1,800 runs.

Below is the comparison with benign priority Raw. This baseline places predicted-benign nodes first, orders each predicted class by Raw uncertainty, and directly reviews the first $H$ nodes without an audit. Table~\ref{tab:benign-priority} compares it with \method\ on the same frozen fits and budgets. \method\ improves both mean gains on these two tasks; the comparison evaluates the complete policies, not the graph score alone.

\begin{table}[!htbp]
\centering
\setlength{\tabcolsep}{6pt}
\caption{Benign-priority comparison. Entries are $G/G_{\mathrm{OOD}}$, relative to the original Raw ranking.}\label{tab:benign-priority}
\begin{tabular}{lrr}\toprule
Task & Benign priority + Raw & \method\\\midrule
X-IIoTID / RDOS & $-102.67 / 0.00$ & $+340.30 / +461.00$\\
UNSW / Fuzzers & $+91.45 / +104.85$ & $+102.95 / +105.72$\\
\bottomrule\end{tabular}
\par\vspace{3pt}\begin{minipage}{\textwidth} Means over three X-IIoTID and 20 UNSW fits. PairAudit averages 600 audit runs per fit; all paid labels count toward $H$.\end{minipage}
\end{table}
\FloatBarrier

\subsection{Observation schedule and implementation}
Follow Section~\ref{sec:audit}. The pair permutation uses the recorded audit seed plus 1,000,000. For nominal cost fraction $f\in\{.05,.1,.2,.3,.5\}$, use
\[
m_f=\min(L-1,H-L,\max(1,\lceil fH/2\rceil)).
\]
Deduplicate and sort these counts. If $L<2$, review $R_H$ without pair queries. Query only previously unseen endpoints and reuse their labels.

For $H=1{,}000$, $\tau=1/3$ and $L=333$, a first look at 25 pairs uses 50 labels. Completion uses 50 audited nodes, 308 nodes from the chosen side and the first 642 nodes of $B$, totaling $H$.

\FloatBarrier

\section{Audit Calculations and Diagnostics}\label{app:proofs}
\subsection{Harm diagnostics and pairing}

For the harmful-authorization diagnostic, let $O=C_{I_m}\cup D_{I_m}$ contain all paid pair endpoints, and let $R^O$ contain the first $H-|O|$ unreviewed nodes in Raw order. An authorization is harmful when $U(S\setminus O)<U(R^O)$. Total workflow loss instead means $G<0$, including investigation costs. Literal Raw continuation $R^O$ need not coincide with the local fallback in Equation~(\ref{eq:completion}); these comparison targets are therefore reported separately.

Every bijective pairing preserves $\sum_j\delta_j=U(C)-U(D)$, but can change ties, sampling variance and stopping time. A common permutation of fixed pairs changes only review order.

\subsection{Exact posterior evaluation}
Under the working model, $X\mid\theta,n\sim\operatorname{Multinomial}(r,\theta)$. Integrating $\theta\mid n\sim\operatorname{Dirichlet}(\alpha_+,\alpha_-,\alpha_0)$ gives, with $\alpha_\cdot=\sum_t\alpha_t$,
\begin{equation*}
\Pr(X=x\mid n)=\frac{r!\,\Gamma(\alpha_\cdot)}{\Gamma(\alpha_\cdot+r)}\prod_{t\in\{+,-,0\}}\frac{\Gamma(\alpha_t+x_t)}{x_t!\,\Gamma(\alpha_t)},\qquad \sum_t x_t=r.
\end{equation*}
Dirichlet aggregation makes $\theta_++\theta_-\sim\operatorname{Beta}(\alpha_++\alpha_-,\alpha_0)$ independent of $\rho=\theta_+/(\theta_++\theta_-)\sim\operatorname{Beta}(\alpha_+,\alpha_-)$, conditional on $n$. Integrating the binomial counts yields Equation~(\ref{eq:predictive}); averaging over $V$ yields Equation~(\ref{eq:predictive-tail}). A cached Polya-urn recursion evaluates the same probability deterministically and is checked against the beta-binomial sum.

\FloatBarrier

\section{Datasets, Graphs and Backbone Qualification}\label{app:data}
\begin{table}[!htbp]
\centering

\setlength{\tabcolsep}{4pt}
\caption{Known-source qualification. Ranges span the fitted configurations, not independent events.}\label{tab:quality}
\begin{tabular}{lrrrrr}\toprule
Task & Fits & Train acc. & Train F1 & Val. acc. & Val. F1\\\midrule
UNSW / Fuzzers & 20 & 0.996--0.997 & 0.993--0.994 & 0.995--0.997 & 0.992--0.995\\
CICIDS / Hulk & 10 & 0.979--0.981 & 0.962--0.965 & 0.979--0.981 & 0.962--0.966\\
X-IIoTID / RDOS & 3 & 0.952--0.953 & 0.951--0.952 & 0.939--0.945 & 0.937--0.944\\
TON / XSS & 3 & 0.985--0.986 & 0.982--0.984 & 0.986--0.986 & 0.982--0.984\\
IoT23 / ARP & 3 & 0.959--0.960 & 0.866--0.868 & 0.958--0.959 & 0.860--0.865\\
IoMT / ARP & 3 & 0.968--0.974 & 0.951--0.961 & 0.964--0.971 & 0.945--0.956\\
BoT / DDoS-TCP & 3 & 0.932--0.950 & 0.929--0.946 & 0.926--0.951 & 0.923--0.948\\
\bottomrule\end{tabular}
\par\vspace{3pt}\begin{minipage}{\textwidth} F1 is macro averaged.\end{minipage}
\end{table}

All reported detectors satisfy known training and validation macro-F1 $\ge.8$. Table~\ref{tab:quality} covers the main-text tasks; Appendix~\ref{app:industrial} reports the industrial extensions. Binary tasks use attack/benign labels: main-text tasks prioritize predicted-benign nodes, while Table~\ref{tab:attack-priority} prioritizes predicted-attack nodes. Multiclass tasks and Physics have no priority.

\subsection{Primary security tasks}
\paragraph{UNSW-NB15 and CICIDS2017.}
Fuzzers (UNSW-NB15) and DoS Hulk (CICIDS2017) are held out from training and validation \citep{moustafa2015unsw,sharafaldin2018toward}. Target pools combine known test nodes and the held-out subtype. We use 20 GraphSAGE and 10 APPNP fits, respectively. Relations use available endpoint, protocol, service/session or temporal fields, never labels. UNSW fits the detector on source-induced edges, then computes posteriors on the unlabelled full graph; source tokens may contain target covariate context. 

\paragraph{X-IIoTID / RDOS.}
Hold out RDOS from X-IIoTID training and validation, retaining attack/benign review labels \citep{alhawawreh2022xiiotid}. Use the author's release and the 55 numeric predictors described in Appendix~\ref{app:industrial}, excluding labels, addresses, ports, timestamps, service and auxiliary detector alerts. Remove conflicting binary-label duplicates and deduplicate predictor vectors. Retain up to 5,000 rows per group at preprocessing, then 5,000 for benign and RDOS groups and 2,000 for each other group. Split known groups 60/20/20 within subtype; all 5,000 RDOS nodes enter the target only. Each fit has 27,541 training, 9,181 validation and 14,187 target nodes.

Fit imputation, signed-log transformation and standardization on training data, then clip to $[-10,10]$. Build separate symmetrized ten-nearest-neighbor source and target feature graphs. Three GraphSAGE fits (seeds 9269301--9269303) use hidden size 64, dropout $.3$, Adam learning rate $.01$, weight decay $.0005$ and inverse-square-root class weights. Known-validation macro-F1 selects checkpoints within 300 epochs, with patience 45 after epoch 100. Audit seeds are 9270000--9270599.

\paragraph{TON-IoT.}
XSS is source-excluded; three GraphSAGE fits retain attack/benign labels \citep{moustafa2021ton}. Separate source and target graphs use source/destination addresses, directed address pairs and protocol/port groups. Features include duration, byte and packet counts, missed bytes, protocol, service and connection state. Subtypes never enter scoring or audit.

\subsection{Stress pools and multiclass review}
\paragraph{IoMT / ARP.}
Hold out ARP Spoofing in CIC IoMT 2024 \citep{ciciomt2024}. Labels are capture-derived.

Remove labels/identifiers and conflicting predictor duplicates. Cap preprocessing at 5,000 nodes per subtype, then retain up to 5,000 benign/held-out nodes and 2,000 per other subtype. Known subtypes use seeded 60/20/20 training/validation/test splits without predictor overlap; held-out nodes are target-only. Fit medians, signed-log transformation and standardization on training data, clip features to $[-10,10]$, and build source and target feature-neighbor graphs separately.

Known validation selects among GCN, GraphSAGE, APPNP and ResidualSAGE, subject to qualification. ResidualSAGE wins all three fits. It joins 128-dimensional two-layer feature and GraphSAGE pathways, with dropout $.1$, Adam learning rate $.003$, weight decay $.0005$ and inverse-square-root class weights. Validation early stopping selects within 600 epochs; other backbones use learning rate $.01$ and at most 300 epochs.

\paragraph{CIC IoT 2023 / MITM-ArpSpoofing.}
Hold out MITM-ArpSpoofing in CIC IoT 2023 \citep{ciciot2023}. Deduplication, caps, 60/20/20 known splits and transforms follow IoMT. Three GraphSAGE fits (seeds 9137101--9137103) use hidden size 64, dropout $.3$, learning rate $.01$, weight decay $.0005$ and inverse-square-root class weights. Each has 40,668 training and 13,556 validation nodes. Known-validation macro-F1 selects checkpoints within 300 epochs, with patience 45 after epoch 100.

Each low-OOD cohort contains all 13,556 known test nodes and 137 uniformly sampled held-out attacks. Seeds 9136000--9136002 give three cohorts per fit. Rebuild target edges using only cohort nodes before frozen inference; omitted attacks provide no context. Source and target graphs remain separate.

\paragraph{Stress construction and denominators.}
For IoMT high-OOD pools, set $N'=\min(N,\lfloor N_{\mathrm{OOD}}/a\rfloor,\lfloor N_{\mathrm{known}}/(1-a)\rfloor)$ with $a=.95$, sample $\operatorname{round}(aN')$ OOD nodes and permute the pool. This retains 1,370 ARP nodes and samples 72 known nodes; three seeds give three cohorts per fit. Unlike IoT23, graph context and scores remain from full-target inference. Each pool uses $H=\lceil.1N'\rceil$. In Table~\ref{tab:extension}, $\rho_{\mathrm{err}}=N'^{-1}\sum_i u_iv_i$ counts OOD errors per target node; prevalence also includes correctly predicted OOD nodes.

\paragraph{BoT-IoT multiclass review.}
Use the public 5\% BoT-IoT subset and native category/subcategory labels \citep{koroniotis2019bot}. Review assigns DDoS, DoS or reconnaissance. Normal and Theft are outside this task because of insufficient representation. Hold out DDoS/TCP, retaining DDoS/UDP and DDoS/HTTP as known. Use 18 measurements of protocol/flags/state, counts, durations and rates; exclude IPs, ports, absolute timestamps, sequence IDs, labels and capture aggregates. Deduplication, conflict removal, caps and training-only transforms follow IoMT. Known 60/20/20 splits yield 7,483 training and 2,495 validation nodes. The target has 7,482 nodes, including 4,987 DDoS/TCP nodes; $H=749$. Source and target use separate symmetrized ten-nearest-neighbor graphs.

Three GCN fits (seeds 9139101--9139103) use hidden size 64, dropout $.3$, Adam learning rate $.01$, weight decay $.0005$ and inverse-square-root class weights. Known-validation macro-F1 selects checkpoints within 300 epochs, with patience 45 after epoch 100. Validation class recall exceeds $.87$. All classification errors have unit utility; no class-priority offset is applied.

The three fits share one capture collection and held-out subtype.

\subsection{Industrial tasks with predicted-attack priority}\label{app:industrial}
X-IIoTID contains industrial control and service activity \citep{alhawawreh2022xiiotid}; WUSTL-IIoT-2021 comes from an industrial testbed \citep{wustl2021dataset}. Use the author's X-IIoTID Kaggle release and the public WUSTL mirror\footnote{Kaggle datasets \texttt{munaalhawawreh/xiiotid-iiot-intrusion-dataset} and \texttt{annaamalaiu/wustl-iiot-2021-dataset}.}. Hold out normal CoAP rows identified by \texttt{Service} in X-IIoTID and normal UDP rows (\texttt{Target}=0, \texttt{Proto}=17) in WUSTL. This is a constructed business-type shift, not an official temporal split.

Remove labels, addresses, ports and timestamps; also remove X-IIoTID's service field and auxiliary detector alerts, and WUSTL's IP IDs and epoch-like \texttt{IdleTime}. Retain 55 and 38 numeric features, respectively. Drop conflicting binary-label duplicates and deduplicate predictor vectors. Deterministic caps retain at most 5,000 rows per group, then 2,000 per known group and 5,000 for the held-out group; discard groups below 100 rows. Known groups use seeded 60/20/20 splits, while the held-out group enters only the target. Imputation, signed-log transformation, standardization and clipping follow IoMT; source and target ten-nearest-neighbor feature graphs are separate.

Fix GraphSAGE (hidden size 64, dropout $.3$), Adam learning rate $.01$, weight decay $.0005$ and inverse-square-root class weights. Known-validation macro-F1 selects checkpoints within 300 epochs, with patience 45 after epoch 100. Confirmation seeds 9268201--9268203 yield mean known training/validation macro-F1 of $.947/.939$ for CoAP and $.996/.997$ for UDP. Their target sizes are 14,187 and 7,091, each with 5,000 held-out normal nodes. Use $H=\lceil.1N\rceil$, $\tau=1/3$, unchanged $.95/.05$ thresholds, and audit seeds 9269000--9269599.

\subsection{Rejection control and source limitations}
\paragraph{Coauthor Physics.}
Use native coauthorship edges, word features and field labels \citep{shchur2018pitfalls}. Field 4 is absent from output classes 0--3. Known classes use 50\% training, 20\% validation and the remainder for testing; all 3,519 field-4 nodes enter the 12,813-node target. Three known-validation-selected GraphSAGE fits use no priority. Every field-4 label is automatically a detector error.

\FloatBarrier

\section{Additional Experiments}
The following controls evaluate fixed settings; they do not select deployment parameters from target gains.
\subsection{Review policy implementations}\label{app:baselines}
Each baseline reviews the first \(H\) nodes in its own ranking. \method\ uses its complete policy, including candidate priorities, Raw retention and paid audits. Combining another ranking with our audit would define a hybrid policy.

Trust Score uses the public default without filtering \citep{jiang2018trust}. Its anomaly score is $-d_{\mathrm{other}}/(d_{\mathrm{pred}}+10^{-12})$, using the second nearest known-training neighbor per class in detector input space. Exact-distance acceleration is checked against the original KDTree.

Energy is $-\log\sum_k\exp(\ell_{ik})$ at temperature one. GNNSafe propagates the logit-derived energy over two steps with coefficient $.5$. GRASP uses its public MSP configuration: eight steps, $\alpha=0$, $\tau_1=5$, $\tau_2=50$ and $\delta=1.001$, receiving unlabelled target indices rather than OOD membership.

CICIDS originally lacked saved weights and logits. Recover all ten APPNP fits with the same splits, seeds and 50-epoch recipe, then use one shared probability/logit bank for all methods, including a known-only refitted PairAudit density. Target class predictions match the original banks; scores and individual runs need not be identical. Energy uses actual logits. Other tasks retain their saved checkpoints.

Recovered checkpoint predictions preserve Raw's first-$H$ set; the maximum Raw-score error is $2.39\times10^{-7}$. An original-KDTree check reproduces the Trust Score comparison.

\FloatBarrier

\subsection{Component and relational controls}\label{app:ablations}
\begin{table}[!htbp]
\centering

\setlength{\tabcolsep}{4pt}
\caption{Complete component and relational control metrics. All settings retain the same budget and ordinary correction utility.}\label{tab:mechanism-full}
\begin{tabular}{llrrrrr}\toprule
Scenario & Condition & $G$ & $G_{\mathrm{OOD}}$ & Auth. (\%) & $2m/H$ (\%) & Harm (\%)\\\midrule
UNSW / Fuzzers & PairAudit & +102.95 & +105.72 & 100.000 & 5.71 & 0.000\\
UNSW / Fuzzers & Fixed weights & +95.35 & +98.28 & 100.000 & 5.75 & 0.000\\
UNSW / Fuzzers & Node marginal & +97.95 & +100.35 & 100.000 & 5.51 & 0.000\\
UNSW / Fuzzers & Matched shuffle & +95.32 & +97.81 & 100.000 & 5.72 & 0.000\\
UNSW / Fuzzers & Global shuffle & +57.98 & +60.60 & 98.833 & 7.88 & 0.017\\
\midrule
CICIDS / Hulk & PairAudit & +789.79 & +830.80 & 100.000 & 5.01 & 0.000\\
CICIDS / Hulk & Fixed weights & +788.53 & +831.60 & 100.000 & 5.01 & 0.000\\
CICIDS / Hulk & Node marginal & +778.19 & +818.20 & 100.000 & 5.01 & 0.000\\
CICIDS / Hulk & Matched shuffle & +790.77 & +829.60 & 100.000 & 5.01 & 0.000\\
CICIDS / Hulk & Global shuffle & +697.10 & +720.40 & 100.000 & 5.01 & 0.000\\
\midrule
X-IIoTID / RDOS & PairAudit & +340.30 & +461.00 & 100.000 & 5.07 & 0.000\\
X-IIoTID / RDOS & Fixed weights & +338.62 & +457.33 & 100.000 & 5.07 & 0.000\\
X-IIoTID / RDOS & Node marginal & +354.38 & +466.67 & 100.000 & 5.07 & 0.000\\
X-IIoTID / RDOS & Matched shuffle & +217.96 & +341.33 & 100.000 & 5.21 & 0.000\\
X-IIoTID / RDOS & Global shuffle & +333.01 & +470.67 & 100.000 & 5.07 & 0.000\\
\midrule
TON / XSS & PairAudit & +80.01 & +102.02 & 100.000 & 6.92 & 0.000\\
TON / XSS & Fixed weights & +78.16 & +97.66 & 100.000 & 7.02 & 0.000\\
TON / XSS & Node marginal & -39.86 & -16.69 & 51.556 & 30.89 & 20.333\\
TON / XSS & Matched shuffle & +65.36 & +93.78 & 100.000 & 8.00 & 0.000\\
TON / XSS & Global shuffle & +46.17 & +62.71 & 99.778 & 9.48 & 0.667\\
\bottomrule\end{tabular}
\end{table}

Controls use the first 300 audit seeds per fit, matched to PairAudit's 600-run evaluation.
\textbf{Fixed weights} sets $\eta_k=0$ with the same Gaussian components and rebuilt reference ranks. \textbf{Node marginal} scores only node coordinates of the fitted density while retaining the gate. It removes explicit neighbor coordinates, but the frozen GNN and training responsibilities still contain graph information.

\textbf{Matched shuffle} permutes neighbor-token halves within data role, predicted class, confidence quartile and feature cluster. Quartiles use known validation; 32 MiniBatchKMeans clusters use known training features. Refit the density and gate after shuffling. \textbf{Global shuffle} permutes within data role only. Each uses one token permutation and 300 audit runs per fit, not a permutation test. TON's node-marginal control has 20.33\% harmful authorized continuations versus 0.78\% harmful remaining pair sums; the distinction is defined in Appendix~\ref{app:proofs}.

\FloatBarrier

\subsection{Component-count sensitivity}\label{sec:component-sensitivity}
Table~\ref{tab:component-sensitivity} fixes $M_k=M$ and selects gate settings on known validation labels. Each count uses the same 48 fits, 63 workloads and 600 audit seeds per workload. These are fixed-count comparisons, not searches up to $M$.
\begin{table}[!htbp]
\centering

\setlength{\tabcolsep}{4pt}
\caption{Component-count sensitivity: net corrections $G$ at the same total review budget.}\label{tab:component-sensitivity}
\begin{tabular}{lrrrrrrr}\toprule
Scenario & Reported & $M=1$ & $2$ & $3$ & $4$ & $8$ & $16$\\\midrule
UNSW / Fuzzers & 102.95 & 103.99 & 72.32 & 81.27 & 102.95 & 103.83 & 103.62\\
CICIDS / Hulk & 789.79 & 720.16 & 792.47 & 792.94 & 789.60 & 800.41 & 789.80\\
X-IIoTID / RDOS & 340.30 & 334.34 & 335.71 & 337.40 & 340.30 & 344.53 & 347.21\\
TON / XSS & 80.01 & -22.07 & 65.20 & 73.25 & 80.01 & 72.63 & 67.73\\
IoMT / ARP, full & 44.19 & -6.47 & 30.03 & 46.11 & 44.38 & 39.16 & 48.35\\
IoMT / ARP, 95\% OOD & 44.33 & 41.37 & 42.65 & 43.66 & 44.22 & 42.87 & 43.32\\
IoT23 / ARP, 1\% OOD & 58.59 & 19.78 & 51.58 & 62.58 & 58.59 & 60.72 & 58.51\\
BoT / DDoS-TCP & 37.81 & -18.59 & -33.47 & -35.44 & 37.81 & 41.97 & 92.42\\
Coauthor Physics & -22.97 & -20.28 & -21.64 & -23.47 & -22.97 & -19.75 & -22.00\\
\bottomrule\end{tabular}
\par\vspace{3pt}\begin{minipage}{\textwidth} Reported uses known-validation density selection; other columns fix $M_k=M$ for every class.\end{minipage}
\end{table}

Sensitivity varies by task. At $M=1$, net gain is negative on TON and full-pool IoMT; BoT is negative for $M=1,2,3$. Larger counts can help, but are not uniformly better. Deployment counts must be selected from validation density fit, not these test gains.

\FloatBarrier

\subsection{Exchange-fraction sensitivity}\label{app:tau}
\begin{figure}[!htbp]
\centering\includegraphics[width=\textwidth]{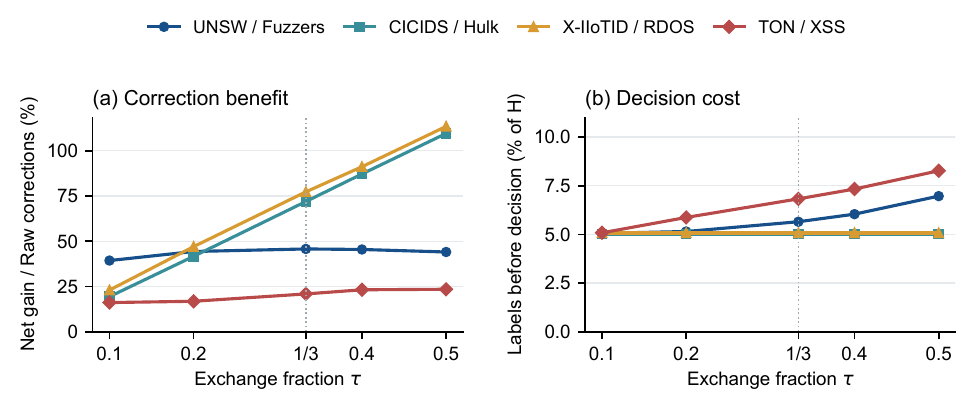}
\caption{Exchange-fraction sensitivity at fixed $H$. (a) Net gain relative to mean Raw corrections. (b) Decision labels as a fraction of $H$. Points average the same primary fits and 300 audit runs per fit; the dotted line marks $\tau=1/3$. Table~\ref{tab:tau} includes exact gains, authorization and harm rates.}\label{fig:tradeoff}
\end{figure}
\begin{table}[!htbp]
\centering

\setlength{\tabcolsep}{4pt}
\caption{Sensitivity to the exchange fraction $\tau$. Each setting has 300 audit runs per fit.}\label{tab:tau}
\begin{tabular}{llrrrrr}\toprule
Scenario & Setting & $G$ & $G_{\mathrm{OOD}}$ & Auth. (\%) & Cost (\%) & Harm (\%)\\\midrule
UNSW / Fuzzers & 0.1 & +88.52 & +88.80 & 100.000 & 5.06 & 0.000\\
UNSW / Fuzzers & 0.2 & +99.78 & +101.16 & 100.000 & 5.15 & 0.000\\
UNSW / Fuzzers & $1/3$ & +102.95 & +105.74 & 100.000 & 5.65 & 0.000\\
UNSW / Fuzzers & 0.4 & +102.30 & +105.66 & 100.000 & 6.04 & 0.000\\
UNSW / Fuzzers & 0.5 & +99.16 & +104.02 & 100.000 & 6.97 & 0.000\\
\midrule
CICIDS / Hulk & 0.1 & +213.15 & +223.90 & 100.000 & 5.01 & 0.000\\
CICIDS / Hulk & 0.2 & +457.60 & +482.80 & 100.000 & 5.01 & 0.000\\
CICIDS / Hulk & $1/3$ & +789.81 & +830.80 & 100.000 & 5.01 & 0.000\\
CICIDS / Hulk & 0.4 & +956.85 & +1000.30 & 100.000 & 5.01 & 0.000\\
CICIDS / Hulk & 0.5 & +1202.68 & +1248.10 & 100.000 & 5.01 & 0.000\\
\midrule
X-IIoTID / RDOS & 0.1 & +101.41 & +129.00 & 100.000 & 5.07 & 0.000\\
X-IIoTID / RDOS & 0.2 & +207.19 & +271.00 & 100.000 & 5.07 & 0.000\\
X-IIoTID / RDOS & $1/3$ & +340.34 & +461.00 & 100.000 & 5.07 & 0.000\\
X-IIoTID / RDOS & 0.4 & +401.48 & +555.00 & 100.000 & 5.07 & 0.000\\
X-IIoTID / RDOS & 0.5 & +499.59 & +696.33 & 100.000 & 5.07 & 0.000\\
\midrule
TON / XSS & 0.1 & +61.89 & +70.67 & 100.000 & 5.08 & 0.000\\
TON / XSS & 0.2 & +64.63 & +83.83 & 100.000 & 5.87 & 0.000\\
TON / XSS & $1/3$ & +80.18 & +102.24 & 100.000 & 6.83 & 0.000\\
TON / XSS & 0.4 & +88.96 & +114.25 & 100.000 & 7.33 & 0.000\\
TON / XSS & 0.5 & +90.08 & +118.79 & 100.000 & 8.27 & 0.000\\
\midrule
Coauthor Physics & 0.1 & -23.35 & -23.27 & 0.000 & 5.15 & 0.000\\
Coauthor Physics & 0.2 & -23.11 & -24.12 & 0.000 & 5.16 & 0.000\\
Coauthor Physics & $1/3$ & -22.94 & -23.17 & 0.000 & 5.15 & 0.000\\
Coauthor Physics & 0.4 & -22.46 & -21.87 & 0.000 & 5.15 & 0.000\\
Coauthor Physics & 0.5 & -19.63 & -18.50 & 0.000 & 5.15 & 0.000\\
\bottomrule\end{tabular}
\end{table}

Test $\tau\in\{.1,.2,1/3,.4,.5\}$ on 36 primary and three Physics fits, with 300 audit runs each (58,500 total). Hold scores and $H$ fixed.

Larger exchanges improve CICIDS, X-IIoTID / RDOS and TON mean gains, while increasing TON's decision cost. The default $1/3$ remains positive on all primary tasks but is not universally optimal. At $\tau=.5$, one of 900 TON runs has negative total gain despite no harmful continuation. Physics always rejects and remains negative relative to Raw.

\FloatBarrier

\subsection{Pair-formation robustness}\label{app:pairing}
\begin{table}[!htbp]
\centering

\setlength{\tabcolsep}{4pt}
\caption{Sensitivity to pair formation. Each setting has 300 audit runs per fit.}\label{tab:pairing}
\begin{tabular}{llrrrrr}\toprule
Scenario & Setting & $G$ & $G_{\mathrm{OOD}}$ & Auth. (\%) & Cost (\%) & Harm (\%)\\\midrule
UNSW / Fuzzers & Ordered & +102.95 & +105.74 & 100.000 & 5.65 & 0.000\\
UNSW / Fuzzers & Shuffle C & +102.96 & +105.74 & 100.000 & 5.64 & 0.000\\
UNSW / Fuzzers & Shuffle D & +102.97 & +105.74 & 100.000 & 5.62 & 0.000\\
UNSW / Fuzzers & Shuffle both & +102.95 & +105.72 & 100.000 & 5.69 & 0.000\\
\midrule
CICIDS / Hulk & Ordered & +789.81 & +830.80 & 100.000 & 5.01 & 0.000\\
CICIDS / Hulk & Shuffle C & +789.81 & +830.80 & 100.000 & 5.01 & 0.000\\
CICIDS / Hulk & Shuffle D & +789.79 & +830.80 & 100.000 & 5.01 & 0.000\\
CICIDS / Hulk & Shuffle both & +789.79 & +830.80 & 100.000 & 5.01 & 0.000\\
\midrule
X-IIoTID / RDOS & Ordered & +340.34 & +461.00 & 100.000 & 5.07 & 0.000\\
X-IIoTID / RDOS & Shuffle C & +340.34 & +461.00 & 100.000 & 5.07 & 0.000\\
X-IIoTID / RDOS & Shuffle D & +340.29 & +461.00 & 100.000 & 5.07 & 0.000\\
X-IIoTID / RDOS & Shuffle both & +340.29 & +461.00 & 100.000 & 5.07 & 0.000\\
\midrule
TON / XSS & Ordered & +80.18 & +102.24 & 100.000 & 6.83 & 0.000\\
TON / XSS & Shuffle C & +79.96 & +102.04 & 100.000 & 7.01 & 0.000\\
TON / XSS & Shuffle D & +79.67 & +101.70 & 100.000 & 7.10 & 0.000\\
TON / XSS & Shuffle both & +79.75 & +101.77 & 100.000 & 7.03 & 0.000\\
\midrule
Coauthor Physics & Ordered & -22.94 & -23.17 & 0.000 & 5.15 & 0.000\\
Coauthor Physics & Shuffle C & -22.91 & -23.16 & 0.000 & 5.15 & 0.000\\
Coauthor Physics & Shuffle D & -22.94 & -23.17 & 0.000 & 5.15 & 0.000\\
Coauthor Physics & Shuffle both & -22.91 & -23.16 & 0.000 & 5.15 & 0.000\\
\bottomrule\end{tabular}
\end{table}

Compare the fixed pairing, shuffled $C$, shuffled $D$, and independently shuffled sides on 39 fits, with 300 audit runs per setting (46,800 total). Keep sets, scores and $H$ fixed. Side permutations depend only on configuration and run IDs, before outcomes are accessed.

Primary-task mean gains change by at most $.52$ corrections. All matched primary runs authorize with no harmful continuation or negative total gain; shuffling slightly raises TON's decision cost. Physics always rejects at the first look.

\FloatBarrier

\section{Workload Results and Reproducibility}\label{app:instances}\label{app:repro}
\subsection{Fits and audit runs}
Table~\ref{tab:instances} lists 63 workloads from 48 frozen fits: 36 primary, three Physics, three BoT, three IoMT full, nine IoMT high-OOD and nine IoT23 low-OOD workloads. Each row averages 600 audit runs; cohort means weight nine workloads equally. IoT23 L1--L3 denote source seeds 9137101--9137103; BoT M1--M3 denote 9139101--9139103. Audit seeds randomize pair order, not events. Saved ledgers verify exact $H$, retention of paid nodes and net/OOD gains.

\subsection{Run-level losses}
The 21,600 primary runs have no negative gain or harmful authorization. IoT23 has one harmful authorization and eight negative-gain runs out of 5,400. Full-pool IoMT has 42/1,800 negative-gain runs and BoT has 21/1,800, with no harmful authorization in either. High-OOD IoMT has neither outcome in 5,400 runs. Physics rejects all 1,800 times but loses 22.97 corrections on average through investigation costs.

\clearpage
\begingroup
\renewcommand{\arraystretch}{0.94}
\setlength{\tabcolsep}{3pt}\begin{longtable}{@{}lllrrrrrr@{}}
\caption{Main-text PairAudit configurations and workloads. Each row contains 600 audit runs; fit IDs identify frozen configurations in the supplement.}\label{tab:instances}\\\toprule
Task & Fit ID & Mix/cohort & $N$ & $H$ & $G$ & $G_{\mathrm{OOD}}$ & Auth. (\%) & Cost (\%)\\\midrule\endfirsthead
\multicolumn{9}{l}{Table \thetable\ continued}\\\toprule Task & Fit ID & Mix/cohort & $N$ & $H$ & $G$ & $G_{\mathrm{OOD}}$ & Auth. (\%) & Cost (\%)\\\midrule\endhead
\midrule\multicolumn{9}{r}{Continued on next page}\endfoot\bottomrule\endlastfoot
UNSW / Fuzzers & 21 & Full & 25,305 & 2,531 & +104.96 & +104.97 & 100.00 & 5.11\\
UNSW / Fuzzers & 22 & Full & 25,305 & 2,531 & +166.39 & +171.00 & 100.00 & 5.07\\
UNSW / Fuzzers & 23 & Full & 25,305 & 2,531 & +153.07 & +153.00 & 100.00 & 5.06\\
UNSW / Fuzzers & 24 & Full & 25,305 & 2,531 & +44.11 & +49.32 & 100.00 & 9.27\\
UNSW / Fuzzers & 25 & Full & 25,305 & 2,531 & +96.33 & +98.95 & 100.00 & 5.32\\
UNSW / Fuzzers & 26 & Full & 25,305 & 2,531 & +96.35 & +101.84 & 100.00 & 5.48\\
UNSW / Fuzzers & 27 & Full & 25,305 & 2,531 & +61.83 & +69.37 & 100.00 & 7.49\\
UNSW / Fuzzers & 28 & Full & 25,305 & 2,531 & +111.07 & +110.99 & 100.00 & 5.07\\
UNSW / Fuzzers & 29 & Full & 25,305 & 2,531 & +105.33 & +109.95 & 100.00 & 5.32\\
UNSW / Fuzzers & 30 & Full & 25,305 & 2,531 & +79.99 & +80.91 & 100.00 & 5.28\\
UNSW / Fuzzers & 31 & Full & 25,305 & 2,531 & +61.66 & +62.58 & 100.00 & 5.78\\
UNSW / Fuzzers & 32 & Full & 25,305 & 2,531 & +62.25 & +64.96 & 100.00 & 6.35\\
UNSW / Fuzzers & 33 & Full & 25,305 & 2,531 & +210.23 & +213.00 & 100.00 & 5.06\\
UNSW / Fuzzers & 34 & Full & 25,305 & 2,531 & +63.86 & +65.68 & 100.00 & 5.99\\
UNSW / Fuzzers & 35 & Full & 25,305 & 2,531 & +109.99 & +107.99 & 100.00 & 5.07\\
UNSW / Fuzzers & 36 & Full & 25,305 & 2,531 & +66.85 & +71.48 & 100.00 & 6.40\\
UNSW / Fuzzers & 37 & Full & 25,305 & 2,531 & +160.24 & +162.00 & 100.00 & 5.06\\
UNSW / Fuzzers & 38 & Full & 25,305 & 2,531 & +108.06 & +111.75 & 100.00 & 5.37\\
UNSW / Fuzzers & 39 & Full & 25,305 & 2,531 & +97.26 & +102.80 & 100.00 & 5.39\\
UNSW / Fuzzers & 40 & Full & 25,305 & 2,531 & +99.10 & +101.97 & 100.00 & 5.20\\
CICIDS / Hulk & 51 & Full & 27,114 & 2,712 & +797.03 & +836.00 & 100.00 & 5.01\\
CICIDS / Hulk & 52 & Full & 27,114 & 2,712 & +766.96 & +803.00 & 100.00 & 5.01\\
CICIDS / Hulk & 53 & Full & 27,114 & 2,712 & +793.51 & +841.00 & 100.00 & 5.01\\
CICIDS / Hulk & 54 & Full & 27,114 & 2,712 & +792.09 & +840.00 & 100.00 & 5.01\\
CICIDS / Hulk & 55 & Full & 27,114 & 2,712 & +813.44 & +853.00 & 100.00 & 5.01\\
CICIDS / Hulk & 56 & Full & 27,114 & 2,712 & +751.11 & +788.00 & 100.00 & 5.01\\
CICIDS / Hulk & 57 & Full & 27,114 & 2,712 & +765.54 & +819.00 & 100.00 & 5.01\\
CICIDS / Hulk & 58 & Full & 27,114 & 2,712 & +799.15 & +827.00 & 100.00 & 5.01\\
CICIDS / Hulk & 59 & Full & 27,114 & 2,712 & +800.79 & +852.00 & 100.00 & 5.01\\
CICIDS / Hulk & 60 & Full & 27,114 & 2,712 & +818.29 & +849.00 & 100.00 & 5.01\\
X-IIoTID / RDOS & 9269301 & Full & 14,187 & 1,419 & +364.03 & +454.00 & 100.00 & 5.07\\
X-IIoTID / RDOS & 9269302 & Full & 14,187 & 1,419 & +317.96 & +461.00 & 100.00 & 5.07\\
X-IIoTID / RDOS & 9269303 & Full & 14,187 & 1,419 & +338.91 & +468.00 & 100.00 & 5.07\\
TON / XSS & 68 & Full & 11,407 & 1,141 & +91.12 & +109.59 & 100.00 & 6.19\\
TON / XSS & 71 & Full & 11,407 & 1,141 & +72.64 & +96.65 & 100.00 & 7.86\\
TON / XSS & 74 & Full & 11,407 & 1,141 & +76.28 & +99.84 & 100.00 & 6.70\\
IoT23 / ARP & L1 & 1\%/0 & 13,693 & 1,370 & +60.69 & +21.97 & 100.00 & 9.59\\
IoT23 / ARP & L1 & 1\%/1 & 13,693 & 1,370 & +41.90 & +14.92 & 99.83 & 11.62\\
IoT23 / ARP & L1 & 1\%/2 & 13,693 & 1,370 & +51.65 & +12.29 & 100.00 & 10.21\\
IoT23 / ARP & L2 & 1\%/0 & 13,693 & 1,370 & +61.71 & +17.68 & 100.00 & 8.87\\
IoT23 / ARP & L2 & 1\%/1 & 13,693 & 1,370 & +69.18 & +26.90 & 99.83 & 8.99\\
IoT23 / ARP & L2 & 1\%/2 & 13,693 & 1,370 & +71.12 & +23.23 & 100.00 & 9.26\\
IoT23 / ARP & L3 & 1\%/0 & 13,693 & 1,370 & +55.93 & +16.45 & 100.00 & 10.13\\
IoT23 / ARP & L3 & 1\%/1 & 13,693 & 1,370 & +61.23 & +24.99 & 100.00 & 9.55\\
IoT23 / ARP & L3 & 1\%/2 & 13,693 & 1,370 & +53.87 & +21.86 & 100.00 & 10.27\\
IoMT / ARP & 9124101 & Full & 6,554 & 656 & +46.47 & +42.62 & 100.00 & 7.45\\
IoMT / ARP & 9124101 & 95\%/0 & 1,442 & 145 & +45.00 & +45.00 & 100.00 & 5.52\\
IoMT / ARP & 9124101 & 95\%/1 & 1,442 & 145 & +44.00 & +44.00 & 100.00 & 5.52\\
IoMT / ARP & 9124101 & 95\%/2 & 1,442 & 145 & +41.97 & +41.97 & 100.00 & 5.55\\
IoMT / ARP & 9124102 & Full & 6,554 & 656 & +65.72 & +64.32 & 100.00 & 6.44\\
IoMT / ARP & 9124102 & 95\%/0 & 1,442 & 145 & +44.99 & +44.99 & 100.00 & 5.54\\
IoMT / ARP & 9124102 & 95\%/1 & 1,442 & 145 & +46.00 & +46.00 & 100.00 & 5.52\\
IoMT / ARP & 9124102 & 95\%/2 & 1,442 & 145 & +46.00 & +46.00 & 100.00 & 5.52\\
IoMT / ARP & 9124103 & Full & 6,554 & 656 & +20.39 & +26.60 & 100.00 & 11.95\\
IoMT / ARP & 9124103 & 95\%/0 & 1,442 & 145 & +42.99 & +42.99 & 100.00 & 5.58\\
IoMT / ARP & 9124103 & 95\%/1 & 1,442 & 145 & +44.00 & +44.00 & 100.00 & 5.52\\
IoMT / ARP & 9124103 & 95\%/2 & 1,442 & 145 & +44.00 & +44.00 & 100.00 & 5.53\\
BoT / DDoS-TCP & M1 & Full & 7,482 & 749 & +29.00 & +98.06 & 98.50 & 15.46\\
BoT / DDoS-TCP & M2 & Full & 7,482 & 749 & +42.82 & +103.42 & 100.00 & 11.66\\
BoT / DDoS-TCP & M3 & Full & 7,482 & 749 & +41.62 & +103.31 & 100.00 & 12.04\\
Coauthor Physics & 9094101 & Full & 12,813 & 1,282 & -22.10 & -23.00 & 0.00 & 5.15\\
Coauthor Physics & 9094102 & Full & 12,813 & 1,282 & -24.36 & -24.50 & 0.00 & 5.15\\
Coauthor Physics & 9094103 & Full & 12,813 & 1,282 & -22.44 & -22.13 & 0.00 & 5.15\\
\end{longtable}\normalsize

\endgroup

\end{document}